\documentclass[11pt]{article}

\usepackage[preprint]{acl}
\usepackage{times}
\usepackage{latexsym}
\usepackage[T1]{fontenc}
\usepackage[utf8]{inputenc}
\usepackage{microtype}
\usepackage{inconsolata}
\usepackage{graphicx}

\usepackage{todonotes}
\usepackage{booktabs}
\usepackage{soul}
\usepackage{float}
\usepackage{algorithm}
\usepackage{algorithmic}
\usepackage{adjustbox}
\usepackage{multirow}
\usepackage{enumitem}

\title{DaLA: Danish Linguistic Acceptability Evaluation Guided~by~Real~World~Errors}

\author{
 \textbf{Gianluca Barmina\textsuperscript{1}},
 \textbf{Nathalie Carmen Hau Norman\textsuperscript{2}},\\
 \textbf{Peter Schneider-Kamp\textsuperscript{1}},
 \textbf{Lukas Galke Poech\textsuperscript{1}}
\\
\\
 \textsuperscript{1}University of Southern Denmark,
 \textsuperscript{2}University of Copenhagen,
\\
 \small{
   \textbf{Correspondence:} \href{mailto:gbarmina@imada.sdu.dk}{gbarmina@imada.sdu.dk}
 }
}

\begin{document}
\maketitle
\begin{abstract}
We present an enhanced benchmark for evaluating linguistic acceptability in Danish. We first analyze the most common errors found in written Danish. Based on this analysis, we introduce a set of fourteen corruption functions that generate incorrect sentences by systematically introducing errors into existing correct Danish sentences. To ensure the accuracy of these corruptions, we assess their validity using both manual and automatic methods. The results are then used as a benchmark for evaluating Large Language Models on a linguistic acceptability judgement task. Our findings demonstrate that this extension is both broader and more comprehensive than the current state of the art. By incorporating a greater variety of corruption types, our benchmark provides a more rigorous assessment of linguistic acceptability, increasing task difficulty, as evidenced by the lower performance of LLMs on our benchmark compared to existing ones. Our results also suggest that our benchmark has a higher discriminatory power which allows to better distinguish well-performing models from low-performing ones.
\end{abstract}

{\small
Datasets: \href{https://huggingface.co/datasets/giannor/dala}{default\footnote{\href{https://huggingface.co/datasets/giannor/dala}{huggingface.co/datasets/giannor/dala}}}, \href{https://huggingface.co/datasets/giannor/dala_medium}{medium\footnote{\href{https://huggingface.co/datasets/giannor/dala_medium}{huggingface.co/datasets/giannor/dala\_medium}}}, \href{https://huggingface.co/datasets/giannor/dala_large}{large\footnote{\href{https://huggingface.co/datasets/giannor/dala_large}{huggingface.co/datasets/giannor/dala\_large}}}

Code: \href{https://github.com/N-essuno/DaLA}{DaLA GitHub\footnote{\href{https://github.com/N-essuno/DaLA}{github.com/N-essuno/DaLA}}}
}

\begin{figure*}
    \centering
    \includegraphics[width=1\linewidth]{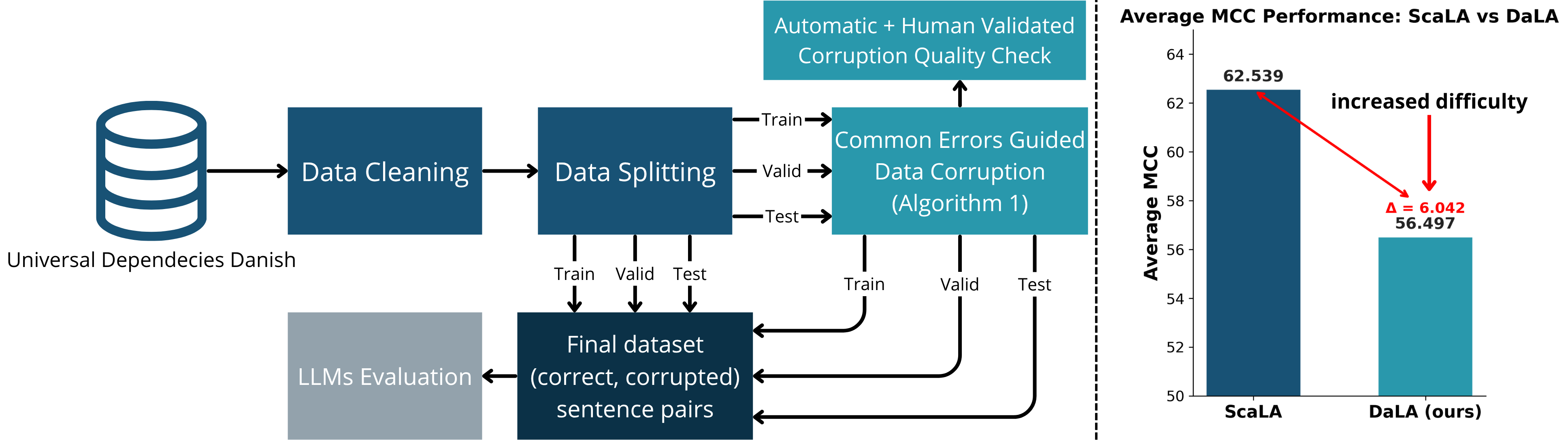}
    \caption{Left: Overview of DaLA creation method, including automatic and human corruption-quality validation, as well as LLM evaluation. Right: Comparison of LLM performance on ScaLA and DaLA, measured with the Matthews Correlation Coefficient (higher is better). LLMs perform worse on DaLA, indicating increased task difficulty.}
    \label{fig:dala-overview}
\end{figure*}

\section{Introduction}

\noindent
Judging linguistic acceptability involves evaluating the well-formedness, nativeness, or naturalness of linguistic forms, such as sentences, typically regarded as a binary classification task. 
Linguistic acceptability has had several applications in the past \citep{myers_acceptability_2017}, including: assessment of the learning of second-language learners \citep{sorace_l2_la}, justification of grammatical analyses \citep{sprouse_2018_acceptability}, study of language development in cognitive processes \citep{mckercher_cognitive}.
Moreover, in recent years, several datasets and benchmarks for linguistic acceptability have been developed to assess the natural language understanding capabilities of large language models (LLMs), some of which we briefly review later.

A model's success in this task suggests that it is able to capture the underlying structural rules of language, indicating deeper linguistic capabilities. However, existing linguistic acceptability resources have substantially been developed for a limited number of widely spoken languages. There is less research and fewer datasets available for low-resource languages, such as Danish.

This scarcity hinders the training and development of language models, particularly multilingual ones. Furthermore, the responses of multilingual models can be influenced by multiple languages, particularly when the languages are closely related (e.g., Danish and Norwegian), which may lead to errors that, while sometimes subtle, remain salient and relevant for human evaluation.

The most relevant prior work on Danish is ScaLA \citep{nielsen_scandeval_2023}. This dataset begins with existing Danish sentences, introduces errors, and creates pairs of sentences, one correct and one incorrect. However, this approach has some limitations: it includes only two corruptions (word removal and word swap); the introduced corruptions are not representative of errors typically made in Danish and are not based on language rules; despite strategies to avoid creating a corrupted sentence that is still grammatically correct, there is no data regarding the accuracy of the corruptions applied; and, finally, the method is restricted to the Universal Dependencies dataset (\citealp{kromann_ddt}; \citealp{johannsen_udd}; \citealp{keson_parole}) (or datasets with the same format), as it relies on POS tags annotations from that dataset.

In this work, we address each of these limitations, introducing a new linguistic acceptability dataset for Danish: DaLA. We introduce 14 new types of corruptions; these are designed to reflect common errors made by Danish speakers; we evaluate the accuracy of the corruptions using both automatic and manual methods; and, finally, we do not require any annotations and make no assumptions about the initial data format, other than the sentence being in Danish and acceptable. An overview of this approach is shown in Figure \ref{fig:dala-overview}. 

In order to compare both methods during the model evaluation phase, we follow the same train, validation, and test proportions of previous work by \citet{nielsen_scandeval_2023}. Hence, the DaLA dataset consists of 1,664 sentence pairs, with 512 for training, 1,024 for validation, and 128 for testing, for a total of 3,328 samples. Each pair includes a grammatically correct sentence and its systematically corrupted counterpart. Bigger versions of this dataset are also made available, up to a total of 7,656 samples.

To facilitate DaLA, we introduce multiple types of corruption, providing both a binary acceptability label and a classification of the specific error type. Each type of corruption is generated using predefined functions that automatically modify the original sentences. This same method can be applied to new sentences enabling a straight-forward expansion of the dataset. Moreover, it can be easily extended by adding new corruption types. To ensure the corruption actually produces a grammatically incorrect sentence, we implemented measures to minimize the cases where a sentence remains unintentionally uncorrupted. Additionally, we individually evaluate each function, that would potentially lead to unintentionally uncorrupted sentences, to assess its effectiveness in introducing errors and maintaining data integrity. For this, as detailed in Section \ref{sec:dala-evaluation}, we employ both automatic and human evaluation. Finally, we evaluate several of the best performing open-source and open-weights large language models on the DaLA corpus and compare their performance on DaLA with the currently most widely used Danish linguistic acceptability corpus, ScaLA \citep{nielsen_scandeval_2023}.

In sum, the key contributions of this work are as follows:
\begin{itemize}[nosep]
    \item We study common errors in Danish and derive 14 corruption functions, applicable to sentences from arbitrary sources datasets.
    \item We create DaLA, a dataset for Danish linguistic acceptability representative of real-world common errors, consisting of 3,328 sentences (extended versions are also available), through applying these corruption functions to Universal Dependencies Data (\citealp{kromann_ddt}; \citealp{johannsen_udd}; \citealp{keson_parole}).
    \item We evaluate the quality of the corruption functions with automatic and manual methods.
    \item We supply baseline results for some of the best models on previous linguistic acceptability datasets in Danish, and show that DaLA is on average 6.04\% harder, going up to 14.55\% for some models. The results suggest also higher discriminatory power between under-performing and high-performing models.
\end{itemize}

\section{Related Work}

\subsection{Linguistic Acceptability}
Early discussions on Linguistic Acceptability primarily focused on grammatical aspects, emphasizing the significance of such judgments from the outset (\citealp{chomsky_57_syntax}; \citealp{chomsky_65_syntax_2}). Over time, the need for more innovative methods to assess linguistic acceptability became evident, expanding the range of available approaches \citep{bard_magnitude}. The introduction of these new methods led to efforts to systematically review and evaluate their effectiveness \citep{sprouse_2013_formal_informal}. Subsequent research delved deeper into these judgments, examining various methodologies and reinforcing their importance despite the inherent complexity of acceptability assessments \citep{schütze_2016_empirical_base}. With the advent of big data and recent technological advancements, machine learning techniques and neural networks have increasingly been explored for predicting linguistic acceptability judgments (\citealp{wagner_2013_grammaticality_judge}; \citealp{lawrence_grammatical_inferenze_2000}).

More recent discussions have shifted the focus from purely grammatical and syntactic factors to the role of cognitive processes, perception, and an individual's familiarity with a given sentence \citep{myers_acceptability_2017}. Recent studies have also investigated the influence of structural complexity and frequency on linguistic acceptability \citep{christensen_complexity_2024}, particularly in Danish, which is of primary interest in this context. Findings from these studies suggest that linguistic acceptability can be shaped by the frequency and everyday use of words and phrases. However, this influence is not absolute. That is, while usage patterns may affect judgments, there is not always a direct correlation between them. Instead, a stronger relationship exists between acceptability judgments and grammatical rules, which appear to have a more consistent impact. In other words, well-established grammatical and structural rules remain among the most reliable predictors of linguistic acceptability judgments. Further evidence for this comes from research on language acquisition, which demonstrates that language learners tend to focus more on sentence structure than on less formal factors such as word frequency or usage patterns when acquiring a language \citep{culbertson_learners_privilege_structures}.

\subsection{Existing Resources}
\label{sec:existing_resources}
Regarding linguistic acceptability, two main types of resources can be considered: those aligned with classical Linguistic Acceptability, which consist of sentence sets annotated with binary judgments (e.g. CoLA \citep{warstadt-etal-2019-neural-cola}), and those following a syntactic minimal pair approach, where sentence pairs are provided one acceptable and the other differing by a single error that renders it unacceptable (e.g. BLiMP \citep{warstadt_blimp_2020}).

\textbf{CoLA} (\citealp{warstadt-etal-2019-neural-cola}; \citealp{cola_lr}) consists of a collection of English linguistic acceptability examples, each annotated as either acceptable or unacceptable. These examples are primarily sourced from scientific linguistic resources, such as books and journal articles. The dataset is divided into two subsets: "in-domain" and "out-of-domain". The in-domain subset, after appropriate splits, is used for both training and testing, whereas the out-of-domain subset (disjoint from the in-domain set) is used exclusively for testing. 

\textbf{NoCoLA} (\citealp{jentoft_nocola_2023}; \citealp{nocola_lr}) consists of two subsets: one following a binary classification approach for linguistic acceptability, and another following a pairwise approach, where each pair contains one acceptable and one unacceptable sentence. All sentences in NoCoLA are sourced from the ASK Corpus (\citealp{tenfjord-etal-2006-ask}; \citealp{ask_lr}), which includes both correct and incorrect sentences.

\textbf{ScaLA} \citep{nielsen_scandeval_2023} provides datasets for multiple languages, including Danish. The original sentences are derived from the respective Universal Dependencies Dataset and are assumed to be linguistically acceptable. Each sentence undergoes one of two types of automatic corruption to generate an unacceptable counterpart.

\textbf{JCoLA} (\citealp{someya_jcola_2024}; \citealp{jcola_lr}) and \textbf{ItaCoLA} (\citealp{trotta_itacola_2021}; \citealp{itacola_lr}) follow a structure closely resembling CoLA, maintaining similar sources and pre-existing acceptability judgments. A notable difference in JCoLA is that more complex examples are included exclusively in the out-of-domain test set, enabling an assessment of a model's generalization ability after training on simpler or more generic cases.

\textbf{CoLAC} (\citealp{hu_revisiting_2023_colac}; \citealp{colac_lr}) similarly follows CoLA's structure but does not differentiate between in-domain and out-of-domain subsets. Instead of relying on acceptability judgments from linguists and academic sources, CoLAC employs judgments from native speakers, arguing that this approach provides a more reliable and usage-based representation of linguistic acceptability.

\textbf{MELA} (\citealp{zhang_mela_2024}; \citealp{mela_lr}) covers multiple languages. It incorporates existing linguistic acceptability resources while creating new datasets for some languages that previously lacked them, namely German, French, Spanish, Arabic, Icelandic, and Japanese. The methodology closely follows that of CoLA, sourcing sentences primarily from books in the fields of linguistics and syntax. 

\textbf{BLiMP} (\citealp{warstadt_blimp_2020}; \citealp{blimp_lr}) adopts a different approach, generating sentences automatically using grammar templates designed by linguists to model common syntactic structures in English. Given templates for acceptable sentences, the approach also generates corresponding unacceptable sentences, enabling controlled dataset construction. 

\textbf{JBLiMP} (\citealp{someya_jblimp_2023}; \citealp{jblimp_lr}), for Japanese, and \textbf{BLiMP-NL} (\citealp{blimp-nl}; \citealp{blimp_nl_lr}), for Dutch, follow a structure similar to BLiMP. However, unlike BLiMP, the former sources sentences from books and scientific articles on syntax and linguistics, incorporating pre-existing acceptability judgments, whereas the latter employs a hybrid approach, manually creating 10 minimal pairs per grammatical paradigm and automatically generating the remaining 90 pairs by prompting a large language model.

\textbf{MultiBLiMP} (\citealp{multiblimp}; \citealp{multiblimp_lr}), is a multilingual resource covering 101 languages, including Danish. It introduces two types of subject-verb agreement errors for creating linguistic minimal pairs through an automated pipeline.

\textbf{DaLAJ} (\citealp{volodina2021dalaj}; \citealp{volodina2023dalaj}; \citealp{dalaj_1_lr}; \citealp{dalaj_2_lr}) for Swedish is based on the processing of existing datasets of sentences produced by second language learners, annotated with errors, and correct sentences extracted from Swedish language teaching books. The resulting dataset comprises an almost equal number of correct and incorrect sentences, with each incorrect sentence containing a single error.

\subsection{Comparison with Related Resources}

\textbf{DaLA} occupies an intermediate position between CoLA-like and BLiMP-like resources. Unlike CoLA and similarly to BLiMP, it introduces automatic corruption, allowing it to be applied to a broader set of sentences even in the absence of explicit acceptability judgments. This enables the creation of a balanced dataset, provided that the sentences are sourced from reliable resources and can be assumed to be acceptable. Similarly to BLiMP, DaLA allows the generation of minimal pair sets by applying a single corruption function to each sentence. However, it has been prepared to be easily extendable and permit multiple corruptions to be applied to the same sentence, further increasing the variability of the dataset. In contrast to BLiMP, and similarly to CoLA, DaLA does not generate sentences automatically from pre-defined patterns but operates on pre-existing sentences. Unlike DaLAJ \citep{volodina2021dalaj}, DaLA includes an automatic and expandable sentence corruption process based on an analysis of the most common errors and therefore does not require a corpus of annotated sentences. Unlike MultiBLiMP \citep{multiblimp}, DaLA introduces 12 additional types of corruption covering a range of grammatical phenomena, compared to the single one addressed in MultiBLiMP (subject-verb agreement). Moreover, the results of LLM evaluations reported on the Danish part of MultiBLiMP show 100\% accuracy for 9 out of 11 models, suggesting that the task is not particularly challenging. Finally, while MultiBLiMP relies on annotations from Universal Dependencies \cite{nivre2016universal, nivre2020universal, de2021universal} and UniMorph \citep{batsuren2022unimorph}, DaLA does not require such annotations to be provided, as they are instead derived automatically. In contrast to all previous approaches DaLA relies on functions derived from real-world errors applicable to arbitrary Danish sentences, making it more representative of real-world scenarios than other resources relying on acceptability judgements derived from scientific sources.

\section{DaLA: Danish Corpus of Linguistic Acceptability}
The total number of corrupted sentences is 3,828. The final dataset therefore consists of up to 7,656 sentences, as each corrupted sentence is paired with its corresponding acceptable version. However, in order to compare DaLA with ScaLA during model evaluation (Section \ref{sec:llm-evaluation}), we keep the same proportions used in ScaLA, reducing the number of samples for each split. We provide also the extended versions using all the 7,656 sentences.

\subsection{Starting Dataset}
The dataset we propose was constructed from sentences in the Universal Dependencies Danish (UDD) corpus (\citealp{kromann_ddt}; \citealp{johannsen_udd}; \citealp{keson_parole}; \citealp{udd_lr}). The sentences in this resource are extracted from original Danish texts; therefore, we assume they are correct and linguistically acceptable, consistent with the approach of \citet{nielsen_scandeval_2023}. Following \citet{nielsen_scandeval_2023}, the dataset was cleaned by removing duplicates, excluding sentences that are too short (<5 tokens), removing overly simple sentences (<5 POS tags), fixing lower and upper bounds on the number of characters per sentence (between 2 and 5,000) and filtering out sentences with ambiguous punctuation.

\subsection{Corruption Type Selection}
Previous studies have aimed to create linguistic acceptability datasets based on examples provided by linguistics resources, existing annotated datasets of correct and incorrect sentences or general linguistic patterns. In contrast, DaLA adopts an approach driven by automatic corruptions of existing sentences and mistakes commonly made by humans. This method aims to generate acceptability samples that better reflect real-world scenarios, thereby providing a language model evaluation that closely aligns with practical use cases. Our selection is based on very common and most noticeable errors made by native Danish speakers: the error selection was informed by linguistic expert judgments and established resources. In particular, we used the list \textit{Typiske Problemer}\footnote{\url{www.sproget.dk/typiske-problemer/}} (common problems) which has the purpose of addressing the most common needs for advice on the Danish language. The list is jointly written by experts from The Danish Language Council and the Society of Danish Language and Literature based on the official spelling rules and other works on the Danish language and grammar. We manually examined each common error and gave them a score from 0 to 5 based on implementation complexity (how difficult is it to automatically corrupt), error type (grammatical, semantic, writing mechanic), and how likely a corruption would produce an incorrect sentence (or how often a corruption is conflated with an acceptable form). After implementing the highest ranking errors, we used an additional criteria of the frequency of applying the corruption to our starting dataset. For instance, the corruption of switching the adverbs \textit{af} (to, along) and \textit{ad} (of, with) would only yield 18 instances. The lowest-ranking problems were almost exclusively informational guides on basic grammar terminology such as what constitutes a word class, sentence, or subordinate clause. We also saw a group of errors related to writing mechanics, e.g. when to use lower or upper case, how to split words at line breaks, and how to use punctuation. Since we focus primarily on grammatical errors, as they are among the most frequent and accurate predictors of linguistic acceptability judgments \citep{christensen_complexity_2024}, we gave this group a lower ranking. In total, 14 corruption types (and corresponding errors) were considered most relevant for implementation alongside with a basic generic corruption derived from \citet{nielsen_scandeval_2023}, resulting in 15 implemented corruptions overall, 14 of which have been introduced by us. For more details on each specific corruption, we refer to Appendix \ref{app:corruption_details}

\subsection{Introduced Corruptions and Implementation}
\label{sec:corruptions}
For each selected error, a dedicated function was developed to inject the error into an input sentence\footnote{The main library used for this purpose was SpaCy \cite{honnibal2020spacy}}. The main focus during implementation was to minimize the generation of corrupted sentences that remain grammatically correct, thereby ensuring the highest possible quality and precision. To achieve this, we adopted a conservative approach: for each corruption, an expert analyzed in which cases its application would most likely result in sentences that were still grammatically correct. In such cases, we preferred applying stricter filtering to the pool of corruptible sentences rather than risk introducing a large number of still-grammatical sentences. Through this conservative strategy, we achieved a balanced representation of each error, as further detailed in Section \ref{sec:corruption-diversity}. The types of errors introduced are:

\begin{itemize}[nosep]
    \item Corruption of indefinite determiners.
    \item Corruption of determiners that can be appended to the end of a noun as a suffix.
    \item Confusion between the pronouns \textit{nogle} and \textit{nogen}.
    \item Confusion between the suffixes \textit{-ende} and \textit{-ene}.
    \item Corruption of subject and object pronouns.
    \item Corruption of the pronouns \textit{som} and \textit{der}.
    \item Corruption of personal and impersonal pronouns (\textit{han/hun/det}).
    \item Commonly observed spelling errors.
    \item Confusion between the verbs \textit{ligge} and \textit{lægge}.
    \item Introduction of the ``R problem'' for verbs, nouns, and adjectives, related to confusion over certain types of words containing suffixes with the letter "r".
    \item Corruption of different types of genitives.
    \item Corruption of the verb \textit{får} and the conjunction \textit{for}.
    \item The two basic corruptions implemented in \citet{nielsen_scandeval_2023}.
\end{itemize}

\subsection{Iterative Corruption Algorithm}
\label{sec:corruption-merge-algo}
The individual corruption functions were combined to produce an algorithm applied to existing Danish sentences for generating the final dataset. Following the method introduced by BLiMP \citep{warstadt_blimp_2020}, a single error is introduced per sentence. The errors considered may vary in representation depending on the distribution of the starting sentence dataset. For example, if the starting dataset contains few sentences with pronouns, errors applicable to pronouns may be less represented. To mitigate this issue, our proposed algorithm (Algorithm \ref{alg:corrupt-dala}) applies corruptions in ascending order according to their frequency of applicability in the starting dataset. This means that the corruption applicable to the fewest sentences is applied first, corrupting all possible sentences, followed by subsequent corruptions in order.

\begin{algorithm}
\footnotesize
\caption{DaLA Iterative Corruption}
\label{alg:corrupt-dala}
\begin{algorithmic}[1]
\STATE \textbf{Input:} List of corruption functions $cor$ sorted by applicability frequency on starting dataset; Starting sentence dataset $sents$
\STATE \textbf{Output:} List of tuples (\textit{corrupted sentence}, \textit{corruption type}) $res$
\FOR{$s$ in $sents$}
    \IF{$s$ is not corrupted}
        \FOR{$f$ in $cor$}
            \STATE $candidate$ = $f(sent)$
            \IF{$candidate$ is corrupted}
                \STATE Mark $s$ as corrupted
                \STATE Append tuple ($candidate$, error type from $s$) to $res$
            \ENDIF
        \ENDFOR
    \ENDIF
\ENDFOR
\STATE \textbf{return} $res$
\end{algorithmic}
\end{algorithm}

Any change in the dataset or in the set of applied corruptions requires recalculating the applicability frequency to account for a potentially different error distribution or data imbalance. However, this process is quick and inexpensive.

\subsection{Output Dataset}
The final corpus is divided into training, validation, and test splits, and is ready to be used as a finetuning dataset for LLMs in subsequent evaluations. Similar to \citet{warstadt_blimp_2020}, each split consists of minimal pairs of a correct sentence and a corrupted version of it with a single error. For each corrupted sentence the label and corruption type are also indicated. The final dataset sizes are:

\begin{itemize}[nosep]
    \item Training: 512 pairs = 1024 sentences
    \item Validation: 128 pairs = 256 sentences
    \item Test: 1024 pairs = 2048 sentences
\end{itemize}

These dimensions align with those used in the previous ScaLA work \citep{nielsen_scandeval_2023} and have been maintained for the sake of comparability. However, these data represent only a subset of the original potential dataset. Therefore, we present two additional versions of DaLA that utilize all available examples. Below, we report the proportions and the number of examples for the training, validation and test splits, respectively:

\begin{itemize}[nosep]
    \item DaLA Medium
        \begin{itemize}[nosep]
            \item Proportions: 0.6 / 0.05 / 0.35
            \item Samples: 4592 / 386 / 2678
        \end{itemize}
    \item DaLA Large
        \begin{itemize}[nosep]
            \item Proportions: 0.8 / 0.05 / 0.15
            \item Samples: 6124 / 384 / 1148
        \end{itemize}
\end{itemize}

\subsection{Corruption Diversity}
\label{sec:corruption-diversity}
\begin{figure}
    \centering
    \includegraphics[width=1\linewidth]{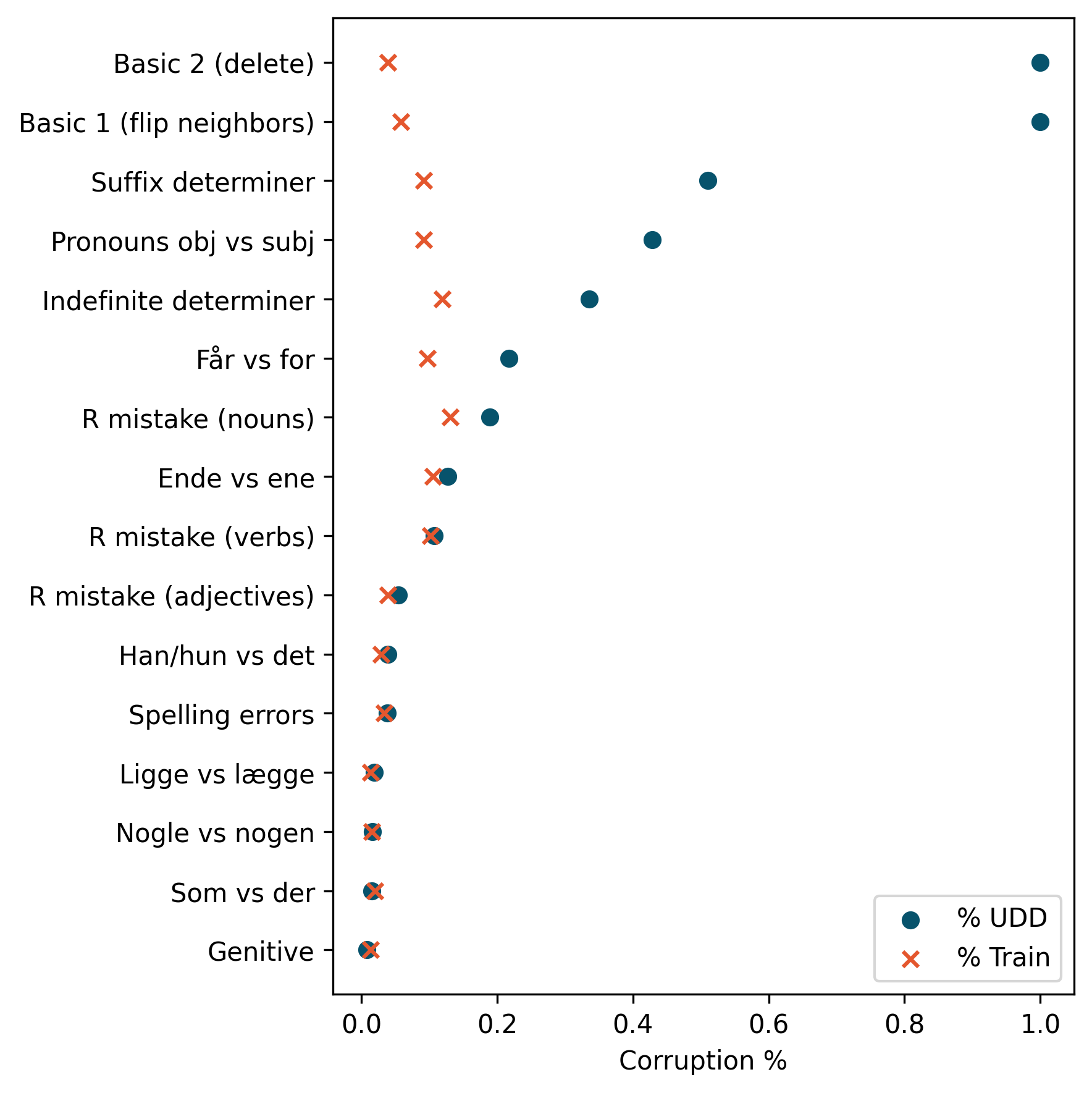}
    \caption{Proportion of corruptible examples among all Universal Dependencies samples (blue circles) vs proportion of actually corrupted examples in training set (red crosses).}
    \label{fig:corruption-proportion}
\end{figure}

From Figure \ref{fig:corruption-proportion}, it can be observed that the ``Basic corruptions'', derived from ScaLA \citep{nielsen_scandeval_2023}, have the potential to cover the entire starting dataset, as they are highly generic. However, they are less representative of common real-world errors. Furthermore, we can see that Algorithm \ref{alg:corrupt-dala} has successfully limited the overrepresentation of specific corruptions, enabling a more balanced distribution of all error types. This approach favors less represented errors while constraining the proportion of ``Basic corruptions'', which, although less significant, remain useful as ``rollback corruptions'' and for improving generalization. Although we prioritized underrepresented errors, it is not possible to avoid the distribution of the original dataset to influence the final proportions. However, the initial dataset was collected with a focus on text diversity, and it may as well be caused by the natural distribution of the various grammatical phenomena. Nonetheless, to prevent excessive underrepresentation, we ensured that less frequent errors are included in the dataset as much as possible. Finally, we verified the distance between the training and test datasets, ensuring a similar representation of the different errors. We observed a Jensen-Shannon (JS) divergence of less than 0.01, full results can be seen in Appendix \ref{app:train-test-distance}. Therefore, the training and test distributions are sufficiently similar and appropriate for use as such.

\section{Validating the Corruption Method}
\label{sec:dala-evaluation}

For all automatic corruptions that could potentially result in a still-acceptable sentence, we conducted a hybrid validation comprising automatic classification and manual validation to ensure high quality of the final dataset. The only corruption excluded from the validation was \textit{spelling errors}, as these were derived from a list of spelling errors identified during a work on finding word candidates for the Danish Dictionary \cite{sorensen2023trawling, appel2024jagten}. Importantly, each spelling error occurs in corpus data and is manually categorized by lexicographers to follow common spelling error patterns for Danish and, therefore, does not represent a random typing mistake. As the primary aim of the method is to generate unacceptable sentences, we use precision as the evaluation metric, defined as the proportion of truly unacceptable sentences (post-corruption) among all corrupted sentences. Thus, a key concern is that the corrupted sentences are in fact unacceptable rather than modified but still acceptable. The validation procedure consisted of two phases: an initial automatic validation phase (see Section \ref{sec:auto-eval}), followed by a manual validation phase by an expert (see Section \ref{sec:manual-eval}).

\subsection{Automatic Validation}
\label{sec:auto-eval}

The automatic system employed for validating corruptions is called \textit{writeassistant}\footnote{\texttt{\textcolor{blue}{\href{www.writeassistant.com}{www.writeassistant.com}}}}. This is a free platform provided by a Danish language technology company. The system accepts Danish text as input and produces grammaticality judgments, identifying which words or parts of the sentence contain errors, the types of errors, explanations, and automatic corrections. According to internal evaluations conducted by the company and made available to us, the system is among the leading grammatical error correction tools for the Danish language. We also performed a further evaluation of the system's precision in identifying truly corrupted sentences. We randomly extracted a sample of sentences (all, if the sentences were few, 50 otherwise) detected as incorrect for each error type and manually verified the correctness of the system's classification. The average precision (proportion of correctly classified incorrect sentences among the total number of corrupted sentences considered across all errors) is 99.7\%. For each corruption type, we analyzed which internal error categories the system detected and subsequently automated this process across all corrupted sentences. This allowed us to obtain a precision score for each corruption. Despite the generally high performance of the system, for certain error types, it showed low recall, meaning that some corrupted and actual incorrect sentences were still classified as correct or were assigned wrong error categories. Subsequent linguistic analysis revealed that many such sentences were indeed erroneous but went undetected by the system, prompting a second round of manual evaluation for these cases.

\subsection{Manual Validation}
\label{sec:manual-eval}
Following the automatic evaluation of each error type, a second manual evaluation was conducted on the corrupted sentences that were either not classified as erroneous or were assigned a different error category by the automatic system. The manual evaluation was performed by a native Danish speaker with experience in the fields of linguistics and lexicography. The task was to check whether the corruption did indeed contain the intended error. In cases of unsuccessful corruption, a comment was made on why the corruption failed. Some of the comments covered examples like error in the original sentence, the corruption conflated with another viable form, and a corruption on a named entity. For all corruptions except two, the manual evaluation covered all false positives, resulting in an accurate update of the precision. The other two corruptions, namely \textit{pronoun corruption} and \textit{suffix determiner corruption}, are relatively common and thus generated a higher number of examples for evaluation compared to the others. Consequently, we annotated a subset of 304 examples: 140 out of 877 examples for \textit{pronoun corruption} and 164 out of 1199 examples for \textit{suffix determiner corruption}.
Keeping in mind that a positive is a sentence which after corruption is grammatically wrong, we then estimate the new precision based on this subset through the following equation: 

\begin{equation} \label{eq:corruption-quality-estimation}
    prec_\mathrm{new} = \frac{tp_\mathrm{new}}{\# \mathrm{corruptions}}
\end{equation}

$$ tp_\mathrm{new} = \frac{tp_\mathrm{man}}{tp_\mathrm{man}+fp_\mathrm{man}} fp_\mathrm{auto} + tp_\mathrm{auto} $$ 

where $tp_\mathrm{new}$ represents the newly estimated total true positives. $fp_\mathrm{auto}$ and $tp_\mathrm{auto}$ refer to the totals of false positives (classified as still correct after corruption) and true positives (classified as grammatically corrupted), respectively, detected by the automatic evaluation. $tp_\mathrm{man}$ and $fp_\mathrm{man}$ correspond to the total number of samples extracted from $fp_\mathrm{auto}$, manually annotated as true positives or false positives, respectively.
 Since during the annotation we observed recurrent linguistic error patterns, we argue that our estimate is reliable.

\subsection{Corruption Quality Results}

The final results of the corruption quality validation are reported in Table \ref{tab:corruption-evaluation}. For each error we report the precision (proportion of post-corruption unacceptable sentences among all corrupted sentences for that error) calculated using the automatic validation method and then adjusted with manual validation according to Equation \ref{eq:corruption-quality-estimation}. The average precision of the corruptions is approximately 0.957, which successfully validates our method for introducing errors into initially correct sentences.

\begin{table}[ht]
\footnotesize
\centering
\begin{tabular}{lcc|cc}
\toprule
\textbf{Corruption Type} & \textbf{Precision} \\
\midrule
Som vs der & 0.983 \\
Ligge vs lægge & 1.0 \\
Nogle vs nogen & 0.968 \\
Genitive corruption & 0.8 \\
Ende vs ene & 0.998 \\
R problem (verbs) & 0.995 \\
Får vs for & 1.0 \\
R problem (nouns) & 0.983 \\
R problem (adjectives) & 1.0 \\
Han/Hun vs det & 0.824 \\
Indefinite determiner & 0.992 \\
Pronouns obj vs subj & 0.935 \\
Suffix determiner & 0.962 \\
\bottomrule
\end{tabular}
\caption{Corruption precision for each error, i.e., the fraction of sentences being corrupted in a way that the output is grammatically incorrect. Evaluation is done semi-automatically with iterative human checks on a subset and subsequent adjustments.}
\label{tab:corruption-evaluation}
\end{table}

\section{Large Language Models Evaluation}
\label{sec:llm-evaluation}
Using the EuroEval evaluation framework \citep{nielsen_scandeval_2023}, we evaluated some of the best-performing large language models on the current Danish linguistic acceptability judgment task, ScaLA. Specifically, we selected two of the highest-performing models for each typology tracked by the benchmark: encoder-only, base generative, instruction-tuned, and reasoning. We focused exclusively on open-source or open-weight models to ensure reproducibility. Given that the best-performing models are generally large and relatively expensive to run, we included two smaller models to provide a broader overview of model performances on DaLA. These models still demonstrate high performance on ScaLA, and in some cases, their performance is very close to that of the bigger models.

\begin{table*}[]
    \centering
    \begin{adjustbox}{max width=\textwidth}
        \begin{tabular}{llcc|cc}
        \toprule
            & & \multicolumn{2}{c|}{\textbf{Danish part of ScaLA}} & \multicolumn{2}{c}{\textbf{DaLA (ours)}} \\
            \cmidrule(lr){3-4} \cmidrule(lr){5-6}
            \textbf{Model Type} & \textbf{Model (Hugging Face id)} & \textbf{MCC} & \textbf{F1} & \textbf{MCC} & \textbf{F1} \\
        \midrule
            \multirow{2}{*}{Small} 
                & sarnikowski/electra-small-discriminator-da-256-cased & 70.37 ± 2.73 & 84.82 ± 1.44 & \textbf{65.36 ± 2.09} & \textbf{81.97 ± 1.20} \\
                & jonfd/electra-small-nordic & 66.25 ± 3.95 & 82.24 ± 2.22 & \textbf{64.54 ± 4.99} & \textbf{80.78 ± 2.78} \\
        \midrule
            \multirow{2}{*}{Encoder} 
                & ai-sweden-models/roberta-large-1160k & 75.04 ± 2.84 & 87.19 ± 1.58 & \textbf{71.50 ± 2.09} & \textbf{85.13 ± 1.07} \\
                & kennethenevoldsen/dfm-sentence-encoder-large & \textbf{70.29 ± 4.63} & \textbf{84.60 ± 2.52} & 74.19 ± 7.17 & 86.18 ± 3.78 \\
        \midrule
            \multirow{2}{*}{Instruction-Finetuned} 
                & google/gemma-3-27b-it & 60.10 ± 1.54 & 79.67 ± 0.84 & \textbf{56.77 ± 1.65} & \textbf{78.14 ± 0.83} \\
                & google/gemma-3-12b-it & 56.74 ± 1.63 & 77.59 ± 0.86 & \textbf{50.66 ± 1.75} & \textbf{75.02 ± 0.89} \\
        \midrule
            \multirow{2}{*}{Base Generative} 
                & google/gemma-3-27b-pt & 62.46 ± 1.60 & 80.97 ± 0.80 & \textbf{59.78 ± 2.03} & \textbf{79.56 ± 1.04} \\
                & google/gemma-3-12b-pt & 57.32 ± 2.05 & 77.28 ± 1.38 & \textbf{53.93 ± 1.90} & \textbf{75.80 ± 1.18} \\
        \midrule
            \multirow{2}{*}{Reasoning} 
                & qwen/qwen3-32b & 60.07 ± 1.74 & 79.10 ± 0.89 & \textbf{45.22 ± 1.72} & \textbf{72.43 ± 0.86} \\
                & qwen/qwen3-14b & 54.50 ± 1.77 & 74.71 ± 0.99 & \textbf{40.71 ± 1.79} & \textbf{69.26 ± 0.97} \\
        \bottomrule
        \end{tabular}
    \end{adjustbox}
    \caption{ScaLA and DaLA benchmark results (Matthews Correlation Coefficient and F1 with confidence intervals). \textbf{Scores are highlighted in bold, where DaLA is below ScaLA}, indicating that DaLA is harder.}
    \label{tab:scala-dala-results}
\end{table*}

\subsection{Evaluation Setting}
According to the EuroEval framework's methodology \citep{nielsen_scandeval_2023}, encoder-only models are fine-tuned on the training set and subsequently evaluated on the test set, whereas for encoder-decoder and decoder-only model types, a few-shot in-context learning approach is applied. After initial experiments showed that the default evaluation setting was not allowing the LLMs to properly adapt to new data, ending in similar performances despite the higher quality data, we adjusted some default parameters of the framework, resulting in a longer but more accurate and reliable evaluation process for both datasets. In the fine-tuning setting, we increased the patience parameter from 2 to 20, while in the few-shot evaluation setting, we increased the number of evaluation runs from 10 to 50. The primary evaluation metric is the Matthews Correlation Coefficient (MCC)\footnote{\texttt{\textcolor{blue}{\href{https://www.wikipedia.org/wiki/Phi_coefficient}{wikipedia.org/wiki/Phi\_coefficient}}}}, which is particularly well-suited for imbalanced classes \citep{mcc_advantages}. It ranges from -100\% to +100\%, with 0\% indicating a random guess. Our secondary metric is the unweighted macro-F1 score, calculated as the average of the F1-scores for each class across all evaluation runs. For more details on the default settings, refer to Appendix \ref{app:evaluation-setting-details} and \citet{nielsen_scandeval_2023}.

\subsection{Results}
The results are presented in terms of MCC performance, full results are available in Table \ref{tab:scala-dala-results}. Nine out of the ten models considered exhibit a performance decrease when evaluated with our DaLA dataset compared to the most widely used current benchmark (ScaLA). The average performance decrease across these models is 6.04\%, ranging from 1.71\% to 14.85\%. These results indicate that our new dataset is generally more challenging for some of the best-performing models for the linguistic acceptability task according to the previous evaluation method (ScaLA). The only model showing a performance increase is the \verb|DFM Sentence Encoder Large|. This may be attributed to the fact that it is the biggest Danish-only model among those considered, providing it with both enough generalization power and more specialized knowledge of the Danish language compared to models trained on multilingual data or models trained only on Danish data but too small to have enough Danish language knowledge (e.g. \verb|Electra Small Discriminator|). This supports the conclusion that our dataset is more representative of typical Danish errors and favors models with deeper knowledge of the language. The worst performance on DaLA was observed in reasoning models, which showed an average decline of 14.32\%. When evaluated with ScaLA, these models performed similarly to other models assessed using the few-shot method (i.e., instruction-finetuned and base generative models). However, under DaLA evaluation, while the other few-shot models showed a comparable decline in performance, the reasoning models experienced a substantially sharper decrease. This phenomenon suggests that our dataset, being more representative and diverse than previous benchmarks, is more discriminative regarding model performance and may better capture differences in linguistic acceptability judgments of language models.

\section{Conclusion}
In this work, we introduced DaLA, a real-world-data driven method for evaluating linguistic acceptability judgments in the Danish language. Our proposed approach, based on observation of real errors and leveraging existing sentences, enables the creation of a dataset that closely reflects real-world scenarios, thereby providing a more reliable, precise and authentic evaluation of large language models. This was accomplished through a combination of linguistic analysis of the most common errors made by Danish speakers and rigorous automatic and manual evaluation of the quality of the introduced corruptions. The results demonstrate that DaLA is overall more challenging for some of the best-performing open-source and open-weight large language models according to the previous state-of-the-art evaluation for Danish. Furthermore, the findings suggest that DaLA possesses greater discriminatory power, allowing for a clearer distinction between truly high-performing and underperforming models during the evaluation process.

\section{Limitations}
Our method application is strongly rooted in Danish linguistic rules. Future work could apply this methodology to other languages by developing corruption functions based on common errors in each language. Another extension would be to broaden the types of errors, including not only grammatical errors but also semantic, pragmatic and aesthetics ones, thus providing broader coverage of linguistic acceptability. We observed that some sentences in the employed starting (UDD) dataset used  may be considered outdated or sound odd, however they were a negligible amount with respect to the dataset size and were filtered out during data cleaning. We have chosen UDD to be able to compare our results with previous work. However, our methodology is applicable to sentences from arbitrary source datasets and does not rely on any annotations.

\section{Ethical Considerations}
We do not see any potential harm that could arise from our work focusing on evaluating grammatical competences of language models in Danish.

\section{Acknowledgments}
This work was supported in part by the Danish Foundation Models\footnote{\href{https://www.foundationmodels.dk/}{www.foundationmodels.dk}} project.

\bibliography{dala}

\appendix

\section{Details on Implemented Corruptions}
\label{app:corruption_details}

\textbf{Indefinite determiner}: This error involves swapping the articles \textit{en} and \textit{et}. The implementation detects nouns in the sentence, identifies their associated article, if present, via dependency labels, and swaps it (from \textit{en} to \textit{et} or vice versa).

\textbf{Determiner as suffix}: When a noun is singular and definite in Danish, the articles \textit{en} and \textit{et} may appear as suffixes attached to the noun, which can also be erroneously swapped. This case is handled by identifying specific singular definite nouns using morphological labels.

\textbf{\textit{Nogle} vs \textit{Nogen}}: These pronouns are pronounced similarly but are used in different contexts and frequently confused in written Danish too. \textit{Nogle} is used with plural nouns, while \textit{nogen} is used with singular nouns. The implementation locates the noun immediately following the pronoun and, based on pronoun-noun agreement, performs the swap. Since inverting \textit{nogle} to \textit{nogen} can sometimes yield grammatically acceptable sentences, particularly in interrogative or negative constructions, we implemented checks to identify and exclude such cases.

\textbf{\textit{-ende} vs \textit{-ene} suffixes}: These suffixes are commonly confused due to "silent d mistakes" in Danish. The suffix \textit{-ene} typically appears in nouns, whereas \textit{-ende} occurs in adjectives and verbs. The implementation identifies and swaps these suffixes accordingly.

\textbf{Subject vs Object pronouns}: Another common issue, especially among second language learners (L2), is confusion between subject and object pronouns. The implementation maps and swaps subject pronouns accordingly. In addition to pronoun identification, dependency relations are used to determine the syntactic position: subject pronouns in subject or conjunct positions and object pronouns in object or oblique nominal positions are swapped appropriately.

\textbf{\textit{Som} vs \textit{Der}}: Both are relative pronouns but used in different syntactic positions, fulfilling different roles. The most frequent error is confusing \textit{der}, which is used in subject position, with \textit{som}, used in object position, partly because \textit{der} is considerably more frequent. Dependency labels are leveraged to detect relevant instances and perform the swap between \textit{som} and \textit{der}.

\textbf{Personal pronouns}: This error consists of swapping the personal pronouns \textit{han/hun} with the impersonal pronoun \textit{det}. The most common and clearly erroneous case is substituting \textit{det} for \textit{han/hun}. When a sentence contains a pronoun without an explicit antecedent, such as sentences extracted from texts, evaluating acceptability can be challenging. Thus, the implementation includes a check for the presence of a potential referent noun.

\textbf{Spelling errors}: Frequently observed spelling mistakes in a corpus from the Danish Dictionary are addressed by replacing words with their commonly occurring incorrect variants.

\textbf{\textit{Ligge} vs \textit{Lægge}}: These verbs are often confused due to their similar meanings: \textit{Ligge} is used when there is no object, whereas \textit{Lægge} is used when an object is present. The implementation maps and swaps correct usages of these verbs accordingly through the use of dependency labels.

\textbf{R problem}: Another common error arises from confusion involving the letter \textit{r} in certain contexts. This occurs because the pronunciation and written forms of the alternatives can be very similar.
\textbf{R problem (verbs)}: For verbs, the issue involves confusing the present tense suffix \textit{-rer} with \textit{-re}, and conversely, confusing infinitive verbs ending in \textit{-re} with those ending in \textit{-rer}. These distinctions are implemented using morphological features and POS tags.
\textbf{R problem (nouns)}: The cases considered include plural definite nouns with the suffix \textit{-erne} confused with \textit{-ene}; plural indefinite nouns with the suffix \textit{-ere} confused with \textit{-er}; and singular definite nouns with the suffix \textit{-eren} confused with \textit{-erne}. These cases are handled via morphological features and POS tags.
\textbf{R problem (adjectives)}: This error consists of confusing adjectives ending with the suffix \textit{-ere} with those ending in \textit{-er}, and vice versa.

\textbf{Genitive}: In Danish, several genitive forms exist depending on specific cases (\textit{'s, s, '}), which can be confused or misused. The corruption implementation detects and swaps these different genitive types. Detection relies on morphological features or POS tags, supplemented by suffix checks when morphological features are unavailable.

\textbf{\textit{Får} vs \textit{For}}: This error involves confusing the present tense of the verb \textit{få} with the conjunction \textit{for}. Although this is perhaps the least significant error among those considered, it remains sufficiently relevant for inclusion.

\textbf{Basic corruptions}: Following \cite{nielsen_scandeval_2023}, two simple and broadly applicable generic corruptions have been retained as fallback mechanisms to enable the corruption of virtually any sentence. These corruptions involve deleting a token from the sentence or swapping two tokens, with safeguards to ensure the resulting sentences are ungrammatical (e.g., corruptions are not applied to adjectives).

\section{Train-Test Distribution Distance} 
\label{app:train-test-distance}

\begin{table}[H]
\footnotesize
\centering
\begin{tabular}{|l|r|}
\hline
\textbf{Metric} & \textbf{Train - Test distance} \\
\hline
KL Divergence (Tr || Te) & 0.018222 \\
KL Divergence (Te || Tr) & 0.018126 \\
JS Divergence & 0.004530 \\
Total Variation Distance & 0.082031 \\
Wasserstein Distance & 0.003418 \\
Hellinger Distance & 0.067354 \\
\hline
\end{tabular}
\caption{Distribution distance metrics between train and test splits}
\label{tab:train-test-distances}
\end{table}

\section{Corruption Proportions}
Table \ref{tab:single-corruption-frequencies} presents the maximum proportion of corruptible sentences for each corruption type among all sentences in the starting dataset after the cleaning process (3,828). Notably, the ``Basic corruptions'', derived from ScaLA \citep{nielsen_scandeval_2023}, have the potential to cover the entire dataset, as they are highly generic, but also less representative of real-world common errors.
Table \ref{tab:train-corruption-proportions} shows the proportions of each error in the total of corrupted sentences in the training set, which consists of 512 out of 1,024 sentences, as half of the dataset contains correct sentences and the other half contains incorrect sentences.

\begin{table}[H]
\footnotesize
\centering
\begin{tabular}{lcc|cc}
\toprule
\textbf{Corruption Type} & \textbf{\% UDD} \\
\midrule
Basic corruptions & 1.000000 \\
Suffix determiner & 0.510449 \\
Pronouns obj vs subj & 0.428422 \\
Indefinite determiner & 0.335946 \\
Får vs for & 0.217085 \\
R problem (nouns) & 0.189107 \\
Ende vs ene & 0.127249 \\
R problem (verbs) & 0.107612 \\
R problem (adjectives) & 0.054118 \\
Han/hun vs det & 0.038674 \\
Spelling errors & 0.037909 \\
Ligge vs lægge & 0.019073 \\
Nogle vs nogen & 0.016469 \\
Som vs der & 0.015147 \\
Genitive & 0.007839 \\
\bottomrule
\end{tabular}
\caption{Frequency of each single corruption among the entire starting dataset of correct sentences}
\label{tab:single-corruption-frequencies}
\end{table}

\begin{table}[H]
\footnotesize
\centering
\begin{tabular}{lcc|cc}
\toprule
\textbf{Corruption Type} & \textbf{Proportion} \\
\midrule
R problem (nouns) & 0.130859 \\
Indefinite determiner & 0.119141 \\
Ende vs ene & 0.105469 \\
R problem (verbs) & 0.101562 \\
Får vs for & 0.097656 \\
Pronouns obj vs subj & 0.091797 \\
Suffix determiner & 0.091797 \\
Basic 1 (flip neighbors) & 0.058594 \\
R problem (adjectives) & 0.039062 \\
Basic 2 (delete) & 0.039062 \\
Spelling errors & 0.033203 \\
Han/hun vs det & 0.029297 \\
Som vs der & 0.019531 \\
Nogle vs nogen & 0.015625 \\
Genitive & 0.013672 \\
Ligge vs lægge & 0.013672 \\
\bottomrule
\end{tabular}
\caption{Corruption distribution on training set}
\label{tab:train-corruption-proportions}
\end{table}

\section{EuroEval Default Evaluation Setting}
\label{app:evaluation-setting-details}

In the case of few-shot evaluation, the number of few-shot examples was set to 12. The label assigned by the model was determined based on the token (containing one of the possible labels) with the highest logit value. For more details on the default settings, we refer the reader to the original evaluation framework paper \citet{nielsen_scandeval_2023}.

\end{document}